\documentclass{ecai}

\usepackage[utf8]{inputenc}
\usepackage{graphicx}
\usepackage{amsmath}
\usepackage{amsthm}
\usepackage{booktabs}

\usepackage{enumitem}
\usepackage{tikz}
\usepackage{tikz-cd}
\usepackage{standalone}
\usepackage{tikz-3dplot}
\usetikzlibrary{patterns,arrows,arrows.meta,calc,shapes,shadows,decorations.pathmorphing,decorations.pathreplacing,automata,shapes.multipart,positioning,shapes.geometric,fit,circuits,trees,shapes.gates.logic.US,fit}
\usetikzlibrary{backgrounds,scopes}

\usepackage[all]{xy}
\usepackage{amsmath}
\usepackage{amssymb,mathtools,amsthm}
\usepackage{thmtools}
\newtheorem{definition}{Definition}

\newcommand{\mM}{\mathcal{M}}
\newcommand{\pP}{\mathcal{P}}
\newcommand{\nN}{\mathcal{N}}
\newcommand{\cC}{\mathcal{C}}

\newcommand{\bR}{\mathbb{R}}
\newcommand{\Astar}{A}

\newcommand{\ca}{\mathsf}
\newcommand{\Ob}[1]{\operatorname{obj} \, #1} %
\newcommand{\Hom}[3]{\operatorname{Hom}_{\,#1}\left[#2,#3\right]} %
\usepackage{orcidlink}
\usepackage[noabbrev]{cleveref}

\begin{document}
\begin{frontmatter}
\author[A]{\fnms{Georgios}~\snm{Bakirtzis}\thanks{Corresponding Author. \orcidlinkf{0000-0003-4992-0193}}}
\author[B]{\fnms{Michail}~\snm{Savvas}}
\author[C]{\fnms{Ruihan}~\snm{Zhao}} 
\author[C]{\fnms{Sandeep}~\snm{Chinchali}} 
\author[C]{\fnms{Ufuk}~\snm{Topcu}} 

\address[A]{Universitat Politècnica de Catalunya}
\address[B]{The University of Iowa}
\address[C]{The University of Texas at Austin}

\title{Reduce, Reuse, Recycle:\\ Categories for~Compositional~Reinforcement Learning}

\begin{abstract}
In reinforcement learning, conducting task composition by forming cohesive, executable sequences from multiple tasks remains challenging. However, the ability to (de)compose tasks is a linchpin in developing robotic systems capable of learning complex behaviors. Yet, compositional reinforcement learning is beset with difficulties, including the high dimensionality of the problem space, scarcity of rewards, and absence of system robustness after task composition. To surmount these challenges, we view task composition through the prism of category theory---a mathematical discipline exploring structures and their compositional relationships. The categorical properties of Markov decision processes untangle complex tasks into manageable sub-tasks, allowing for strategical reduction of dimensionality, facilitating more tractable reward structures, and bolstering system robustness. Experimental results support the categorical theory of reinforcement learning by enabling skill reduction, reuse, and recycling when learning complex robotic arm tasks.
\end{abstract}
\end{frontmatter}

\section{Introduction}

Reinforcement learning (RL) is a powerful tool for sequential decision-making, which is crucial in training robots to execute complex tasks. Nevertheless, the challenge of composing multiple tasks---a prerequisite for creating adaptable, interpretable, and versatile robotic systems---persists. Central to this challenge are issues such as high dimensionality problem spaces and task complexity. These issues manifest in several ways: sparse rewards~\citep{gur:2021}, where positive feedback is infrequent, making learning slower and more complex; a lack of robustness~\citep{skalse:2023}, making systems susceptible to variations in the environment or task; and complexities in sequencing and coordinating sub-tasks~\citep{unhelkar:2019}, particularly when tasks have interdependencies or must perform in a specific order.

A compositional perspective alleviates these challenges inherent in multi-task RL by decomposing complex tasks into simpler sub-tasks~\citep{li:2021}. First, principled decomposition reduces the dimensionality of the problem space. Each sub-task can be learned and optimized separately, resulting in more manageable reward structures and quicker learning times. For instance, sparse rewards become less problematic as each sub-task can be associated with its reward structure, providing more frequent feedback to the learning algorithm. Second, by segregating tasks, failures or variations in one component have a lesser chance of disrupting the entire system, improving overall stability and robustness. If a particular sub-task is performing poorly due to environmental variations, it can be isolated, analyzed, and improved without affecting the performance of other tasks. Third, when tasks are modular, the interactions and dependencies between different tasks can be systematically mapped and managed, simplifying coordinating and sequencing sub-tasks.

Despite progress in compositional RL, the challenge of modular task composition remains a barrier to creating versatile robotic systems. Deciding how to break down a complex task into manageable subtasks relies on domain-specific knowledge. The categorical interpretation of RL problem formulations offers an abstraction that systematically captures tasks' interdependencies and compositional structures, leading to a principled way of discovering optimal task decompositions. Additionally, combining policies from different modules to form a coherent overall policy can lead to conflicts or inefficiencies, especially if the components were developed independently~\cite{mohan:2023}. The categorical semantics of RL specifies how policies are combined, ensuring consistency and coherence in the resulting composite policy and enhancing the robustness of integrated policies. Additionally, proof of the existence of categorical structures equips compositional RL with a symbolic knowledge representation for interfacing and gluing sub-task constructions. In this paper, we validate our theoretical results~\cite{bakirtzis:2022} through computational evidence using increasingly complex robotic tasks.\\[1em]

\noindent
\textbf{Merits of category-theoretic RL:}
\begin{itemize}
    \item  \emph{Reduce}: By representing the dynamics between tasks, such as picking up an object and placing it in a box, the categorical formalism streamlines coordination and sequencing, improving the ability to learn increasingly complex tasks through reduction.

  \item \emph{Reuse}: Through abstraction within the categorical framework, the robot can learn a task, like ``picking up an object,'' and then reuse this knowledge across related tasks modularly.

 \item \emph{Recycle}: The categorical formalism extends this reusability further, allowing for the recycling of learned skills across different contexts. For example, a skill developed for lifting a block can be fine-tuned to lift a soda can.
\end{itemize}

\noindent\textbf{Conventions} \quad Subscripts for Markov decision process (MDP) elements refer to a particular definition instantiation; that is, $S_\nN$ refers to the state space of the MDP $\nN$. %

\section{Preliminaries}

Here we give informal definitions of the category-theoretic structures we use. Consult~\citet{lawvere:2009},~\citet{leinster:2014}, or~\citet{mac:1998} for an in-depth treatment of category theory.
   \begin{definition}[Category]\label{def:category}
   A \emph{category} $\ca{C}$ consists of the following data
   \begin{description}
   \item[objects]  a collection of \emph{objects}, denoted $\Ob{\ca{C}}$

   \item[morphisms] for each pair of objects $A,B$, a collection of \emph{morphisms}
              from $A$ to $B$, denoted $\Hom{\ca{C}}{A}{B}$~\footnote{%
             A morphism $f$ in $\Hom{\ca{C}}{A}{B}$ is usually denoted
              as $A \xrightarrow{f} B$.}
   \item[identity] for each object $A$, a morphism $A \xrightarrow{\mathrm{id}_{A}} A$
   \item[composition] for each $A,B,C$ objects, a \emph{composition} operation
       \begin{equation*}
           \circ_{A,B,C}: \Hom{\ca{C}}{B}{C} \times \Hom{\ca{C}}{A}{B}
                                              \to \Hom{\ca{C}}{A}{C}.
       \end{equation*}
    \end{description}
       To represent a category this data has to also respect the following \emph{relations} for each $A \xrightarrow{f} B$, $B \xrightarrow{g} C$, and $C \xrightarrow{h} D$:
       \[
           \mathrm{id}_{B} \circ f = f \qquad f \circ \mathrm{id}_{A} = f \qquad (h \circ g) \circ f = h \circ (g \circ f),
       \]
       meaning that composing with the identity morphism either
       from the left or the right recovers the morphism itself (unitality)
       and the order of operations when composing morphisms does not matter
       as long as the order of operands is unchanged (associativity).
 \end{definition}

   We will work within $\ca{Set}$, the category
   whose objects are sets and morphisms are functions between them.
   For each set $A$, $\mathrm{id}_{A}$ is the identity function from $A$ to itself, and composition is function composition.
Rather than treating sets as collections of items, the categorical interpretation emphasizes the relationships between sets.

\begin{definition}[Commutative diagrams]\label{def:commutative}
A standard diagrammatic way to express composites is $X\xrightarrow{f}Y\xrightarrow{g}Z$
and equations via commutative diagrams of the following form
\begin{displaymath}
\begin{tikzcd}
X\ar[r,"f"]\ar[dr,"h"'] & Y\ar[d,"g"]\\
& Z
\end{tikzcd}\quad\textrm{ which stands for }g\circ f=h.
\end{displaymath}

A commutative square is, instead,

\begin{displaymath}
\begin{tikzcd}
X\ar[r,"f"]\ar[d,"g"'] & Y\ar[d,"g'"]\\
Z\ar[r,"f'"]& W
\end{tikzcd}\quad\textrm{ which stands for }g'\circ f= f' \circ g.
\end{displaymath}
\end{definition}

A diagram \emph{commutes} when the result of a composition is the same regardless of the path we take from one object to another.

\begin{definition}[Pushout]\label{def:pushout}
    A pushout for morphisms $f \colon C \to A$ and $g \colon C \to B$ is an object $W$ together with morphisms $a\colon A \to W$ and  $b\colon B \to W$ such that the square

\begin{displaymath}
\begin{tikzcd}
C\ar[r,"g"]\ar[d,"f"'] & B\ar[d,"b"]\\
A\ar[r,"a"]& W
\end{tikzcd}
\end{displaymath}
commutes, where the morphisms $a$, $b$ are thought of as ``inclusions,'' such that $a \circ f = b \circ g$, and is universal: for any object $W_o$ with morphisms $a_o \colon A \to W_o$ and $b_o \colon B \to W_o$ such that $ a_o \circ f =  b_o \circ g$, there exist a unique morphism $w\colon W \to W_o $ such that $w \circ a = a_o$ and $w \circ b = b_o$.

\begin{displaymath}
\begin{tikzcd}
C\ar[r,"g"]\ar[d,"f"'] & B\ar[d,"b"] \ar[bend left,ddr,"b_o"]\\
A\ar[r,"a"] \ar[bend right,drr,"a_o"] & W\ar[dashed,dr,"w"]\\
& & W_o.
\end{tikzcd}
\end{displaymath}

Another way to state the above is that $W$ is the colimit (see~\citet[chapter 5]{leinster:2014}) of the diagram 
\begin{displaymath}
\begin{tikzcd}
C\ar[r,"g"]\ar[d,"f"'] & B\\
A& 
\end{tikzcd}
\end{displaymath}

We write $W = A \cup_C B$ for the pushout, with the morphisms $f,g$ implied in the notation.
    
\end{definition}

A simple example of a pushout is the disjoint union $A \coprod B$ of two sets, which is the pushout of the morphisms $\emptyset \to A$ and $\emptyset \to B$.

\section{Compositional Reinforcement Learning via~Morphisms and Subprocesses}

In this section, we restate some definitions and results from our theoretical work for completeness~\cite{bakirtzis:2022}. Task interactivity often relies on the Markov decision process (MDP) structure~\citep{ok:2018}. 
We start by developing a definition for MDPs that is congruent with the traditional MDP representation while containing some subtle generalizations, such as added flexibility of action spaces. Our definition is designed to accommodate uniform action spaces across the entire state space. However, we also cater to applications where varying action spaces in different state space regions are more appropriate, each with distinct semantic meanings. 

For instance, the environment remains relatively stable in a robotic arm example. Conversely, consider a drone maintaining a safe altitude; the action space dynamically adjusts when entering an area where this altitude threshold changes. Traditionally, the literature focuses on surjective morphisms on state and action spaces, aiming to reduce these spaces for efficient learning. Our approach seamlessly incorporates this aspect. However, to emphasize the compositionality feature, it is required also to allow for the expansion of state spaces to encompass task components and incorporate morphisms with mixed characteristics (neither purely injective nor surjective).

\begin{definition}[MDP]\label{def:mdp}
An MDP $\mM =(S, \Astar, \psi, T, R)$ consists of:
\begin{itemize}
    \item The state space $S$, a measurable space with a fixed $\sigma$-algebra.\footnote{In probabilistic models, a $\sigma$-algebra is a collection of sets used to define measurable spaces, which are fundamental in defining event probabilities. A ``fixed $\sigma$-algebra'' implies that we are working within a predetermined framework of measurable sets, which is essential for ensuring that all probabilistic operations are well-defined and comparable across different contexts within our model.}
    \item The state-action space $A$, the total set of actions available at all different states, i.e. elements of $S$.%
    \item A %
    function $\psi \colon A \to S$ that maps an action $a \in \Astar$ to its associated state $s \in S$. Every action in the action space $a \in A$ can be taken at a specific state $s \in S$, and $\psi$ maps the action to that state so that $\psi (a) = s$. Equivalently, the set of actions available at $s$ is the pre-image $\psi^{-1}(s) \subseteq A$.
    \item The information of the transition probabilities is given as a function $T \colon \Astar \to \pP_S$, where $\pP_S$ denotes the space of probability measures on $S$.
    \item The reward function $R \colon A \to \bR$.
\end{itemize}
\end{definition}

The above definition of MDP is congruent with the usual definition seen in, for example, \citet{sutton:2018}. The extra information is necessary to formally examine the compositionality feature, for example, to answer which settings composing two MDPs generate a holistic optimal policy. The resulting theorems apply to the usual definition for any practical RL application. A concrete example of MDPs in use is learning for robotics (section~\ref{sec:ex}).

\begin{definition}[Category of MDPs]
MDPs form a category (definition~\ref{def:category}) $\mathsf{MDP}$ whose morphisms are as follows. Let $\mM_i = (S_i, \Astar_i, \psi_i, T_i, R_i)$, with $i=1,2$, be two MDPs. 

A morphism $m = (f,g) \colon \mM_1 \to \mM_2$ is the data of a measurable function $f \colon S_1 \to S_2$ and a function $g \colon \Astar_1 \to \Astar_2$ satisfying the following compatibility conditions: 

\begin{enumerate}
    \item The diagram
    \begin{align} \label{action-space compat diagram}
        \xymatrix{
        \Astar_1 \ar[r]^-{g} \ar[d]_-{\psi_1} & \Astar_2 \ar[d]^-{\psi_2} \\
        S_1 \ar[r]_-{f} & S_2}
    \end{align}
    is commutative (definition~\ref{def:commutative}).%
\end{enumerate}

\begin{enumerate}
    \item[2.] The diagram 
    \begin{align} \label{action-prob compat diagram}
        \xymatrix{
        \Astar_1 \ar[r]^-{g} \ar[d]_-{T_1} & \Astar_2 \ar[d]^-{T_2} \\
        \pP_{S_1} \ar[r]_-{f_*} & \pP_{S_2}}
    \end{align}
    is commutative, where $f_*$ maps a probability measure $\mu_1 \in \pP_{S_1}$ to its pushforward, meaning $\mu_2 = f_* \mu_1 \in \pP_{S_2}$ under $f$.
\end{enumerate}
\begin{enumerate}
    \item[3.] $R_1 = R_2 \circ g \colon A_1 \to \bR$.
\end{enumerate}
\end{definition}
The constant MDP $\mathsf{pt}$ is the MDP $\mathsf{pt}$ whose state space and action spaces are the one-point set. Every MDP $\mM$ admits a unique, natural morphism $\mM \to \mathsf{pt}$ and $\mathsf{pt}$ is the terminal object in $\mathsf{MDP}$.

The two commutative diagrams above show us when two MDPs are \emph{compatible} in that their interfaces agree. Namely, diagram~\eqref{action-space compat diagram} guarantees that if an action $a_1$ in MDP $\mM_1$ is associated to a state $s_1 \in S_1$, then its image action $a_2=g(a_1)$ under $m$ is associated to the image state $s_2=f(s_1)$. Similarly, diagram~\eqref{action-prob compat diagram} ensures that the transition probability from any state $s_1$ to another state $s_1' \in f^{-1}(s_2')$ under taking action $a_1$ in $\mM_1$ is equal to the transition probability from the state $s_2=f(s_1)$ to $s_2' = f(s_1')$ under action $a_2 = g(a_1)$ in $\mM_2$. The third compatibility condition accounts for the reward in our categorical formulation.%

Intuitively, the category of MDPs represents a way to relate different MDPs to each other through morphisms. A morphism between two MDPs is a pair of functions that consistently map states and actions from one MDP to another.
The two commutative diagrams presented above define what it means for these mappings to be consistent. The first diagram ensures that if an action is associated with a specific state in the first MDP, the action in the second MDP must be related to the corresponding state. The second diagram ensures that transition probabilities between states are preserved under the mapping. In other words, how actions transition from one state to another in the first MDP must correspond to how actions transition between the mapped states in the second MDP. When augmented with the reward function, two MDPs must preserve the relationships between states and actions and the rewards associated with those actions.

\paragraph{Subprocesses}
The definition of morphism correctly captures the notion of a subprocess of an MDP. 
\begin{definition}[Subprocess of MDP]\label{subprocess}
We say that $\mM_1$ is a subprocess of the MDP $\mM_2$ if there exists a morphism $(f,g) \colon \mM_1 \to \mM_2$ such that $f$ and $g$ are injective.
We say that $\mM_1$ is a full subprocess if diagram~\eqref{action-space compat diagram} is cartesian. 
\end{definition}

Since $f$ is injective, we may consider the state space $S_1$ as a subset of $S_2$. Moreover, the condition that diagram~\eqref{action-space compat diagram} is cartesian means that the only available actions on $S_1$ come from  MDP $\mM_1$. Thus, $\mM_1$ being a full subprocess of $\mM_2$ implies that an agent following the MDP $\mM_2$ who finds themself at a state $s_1 \in S_1$ will remain within $S_1$ no matter which action $a_1 \in \Astar_1$ they elect to apply.

Conversely, for an MDP $\mM_2$ and any subset $S_1 \subseteq S_2$ there is a canonical subprocess $\mM_1$ with state space $S_1$, whose action space $\Astar_1$ is defined by
\begin{align} \label{def of A_1}
    \Astar_1 \coloneqq \psi_2^{-1}(S_1) \cap T_2^{-1}( f_\star (\pP_{S_1})).
\end{align}

In fact, $\mM_1$ is uniquely characterized as the maximal such subprocess.

\begin{restatable}{proposition}{subprocessprop}
Any subprocess $\mM_1' \to \mM_2$ with state space $S_1$ factors uniquely through the subprocess $\mM_1 \to \mM_2$.
\end{restatable}

This concept of \emph{factoring} reflects the idea that certain morphisms or relationships break down into simpler or more fundamental parts, and \emph{uniqueness} ensures that this breakdown is done in one specific way. In particular, for MDPs, the above proposition establishes a unique intermediate structure or relationship that connects subprocesses.

\smallskip
\noindent
\textbf{Pushouts: a gluing construction} \quad
The categorical notion of pushout models the gluing of two objects along a third object with morphisms to each. 
Interesting categorical properties usually are \emph{universal}. Universal properties represent specific ideals of behavior within a defined category~\citep{spivak:2014,asperti:1991}.
A pushout's universal property is determined by its being minimal in an appropriate, universal sense. As mentioned above, the simplest example is given in the category of sets by the disjoint union $S_1 \coprod S_2$, which can be viewed as the pushout of the two morphisms $\emptyset \to S_1$ and $\emptyset \to S_2$. 

Suppose that we have two MDPs $\mM_1$ and $\mM_2$ together with a third $\mM_3$, which is expressed as a component of both through morphisms $m_1 \colon \mM_3 \to \mM_1$ and $m_2 \colon \mM_3 \to \mM_2$. The \emph{existence} of a pushout operation in the category $\mathsf{MDP}$ allows us to model the composite task obtained by putting together $\mM_1$ and $\mM_2$ and capture the internal behavior of their common component in a maximally efficient way without introducing extra cost in resources or dimensionality or sacrificing accuracy of the representation. The universal property of pushouts guarantees this as minimal gluings along an overlap. Moreover, if $\mM_3$ is a subprocess, then both $\mM_1$ and $\mM_2$ form subprocesses of the composite $\mM$, as desired.

\begin{restatable}{theorem}{pushout} \label{theorem:pushout}
There exists an MDP $\mM = \mM_1 \cup_{\mM_3} \mM_2$ which is the pushout (definition~\ref{def:pushout}) of the diagram in $\mathsf{MDP}$:
\begin{align*}
    \xymatrix{
    \mM_3 \ar[r]^-{m_1} \ar[d]_-{m_2} & \mM_1 \\
    \mM_2
    }
\end{align*}
\end{restatable}

Gluing behaves well with respect to subprocesses. Theorem~1 is the foundation of MDP compositionality. Pushouts of two MDPs are minimal, universal ways of gluing two MDPs along an overlap in full generality. Their existence is the most desired property of the category MDP in that it guarantees that one can always systematically glue MDPs whenever possible and most efficiently. Without Theorem 1, introducing the category MDP would be of limited value, as one would have to resort to ad hoc constructions, which, after proving Theorem 1, would necessarily have to reduce to the pushout construction anyway given its universality.

\begin{restatable}{proposition}{gluing}\label{prop:gluing}
Suppose that $\mM_3$ is a subprocess of $\mM_1$ and $\mM_2$. Then $\mM_1$ and $\mM_2$ are subprocesses of $\mM_1 \cup_{\mM_3} \mM_2$.
\end{restatable}

The above theoretical framework enables us to make compositionality explicit within RL by providing tools to analyze and represent complex agent relationships through universal constructions---constructions that do not depend on a particular problem or definition of compositional RL but apply to any formulation that uses MDPs.

\section{Compositional Task Completion}\label{sec:ex}

We illustrate the implications of the constructions above in the context of compositional task completion, but this is one possible application. In particular, we derive a denotational language for compositional RL based on the properties of subprocess and pushout.

We employ denotational semantics to provide a rigorous mathematical foundation for modeling RL tasks. Denotational semantics precisely define the meanings of constructs without ambiguity, using mathematical objects. This approach is crucial in RL as it allows us to define the components of learning tasks—such as states, actions, rewards, and transitions—in a clear, consistent, and universally applicable way across different scenarios. By using denotational semantics, we ensure that each component of an RL task is described in terms of its effects and interactions, which facilitates an interpretable composition of complex behaviors. This method contrasts with more operational approaches focusing on the computation process itself. The benefit of denotational semantics in our context is its ability to abstract and generalize problem-solving strategies, making it potentially more manageable to apply them to various tasks. This abstraction is advantageous when dealing with complex decision-making environments, where clarity and consistency in task formulation are crucial to developing robust and scalable solutions.

\subsection{Zig-zag Diagrams}
In this subsection, we restate results from our theoretical work~\cite{bakirtzis:2022}.

For designing compositional tasks, we desire to operationalize using the categorical semantics of RL, which involve accomplishing tasks sequentially. In a general setting, we consider the setup given by, what we term a \emph{zig-zag} diagram of MDPs
    \begin{align} \label{zig-zag}
        \xymatrix@R-=0.3cm@C-=0cm{
        & \nN_0 \ar[dl] \ar[dr] & & \nN_1 \ar[dl] \ar[dr] & & \ldots & & \nN_{n-1} \ar[dr] \ar[dl] \\
        \mM_0 & & \mM_1 & & \mM_2 & \ldots & \mM_{n-1} & & \mM_n
        }
    \end{align}
where for each $i=0, \ldots, n-1$, $\nN_i$ is a subprocess of $\mM_i$ (definition~\ref{subprocess}).
    
The composite MDP associated with the above diagram is the MDP $\cC_n$ defined by the inductive rule
\begin{align*}
    \cC_0  &\coloneqq \mM_0,\\
    \cC_1  &\coloneqq \cC_0 \cup_{\nN_0} \mM_1,\\
    & \qquad \vdots\\
    \cC_n  &\coloneqq \cC_{n-1} \cup_{\nN_{n-1}} \mM_n.
\end{align*}

Each subprocess $\nN_i \to \mM_i$ models the completion of a task in the sense that an agent's goal is to find themselves at a state of $\nN_i$ eventually. 
Once the $i$-th goal is accomplished inside the environment given by $\mM_i$, we allow for the possibility of a changing environment and more options for states and actions to achieve the next goal modeled by the subprocess $\nN_{i+1} \to \mM_{i+1}$.

The composite MDP $\cC_n$ is a single environment capturing all the tasks simultaneously.
\begin{itemize}[leftmargin=2.5em,align=left]
    \item[$\left<?\right>$]  Suppose an agent learned an optimal policy for each MDP $\mM_i$ given the reward function $R_i$ for achieving the $i$-th goal for each $i=0, \dots,n$. Under what conditions do these optimal policies determine optimality for the joint reward on the composite MDP $\cC_n$? 
\end{itemize}

One scenario in which optimality is preserved is when the zig-zag diagram is forward-moving, meaning that $\nN_i$ is a full subprocess of $\mM_i$ and the optimal value function $v_\star (s)$ for any state $s$ in the state space of a component $\mM_i$, considered as a state of $\cC_n$, is \emph{monotonic} for subsequent subprocesses $\mM_{i+1}, \ldots, \mM_n$. Monotonicity here means that the expressions
\begin{align*}
    \sum_{s' \in S_i} T(a)(s') (R_i(a) + \gamma \cdot v_\star^{\cC_n} (s') ) \\
    \sum_{s' \in S_i} T(a)(s') (R_i(a) + \gamma \cdot v_\star^{\cC_{[i,n]}} (s') )
\end{align*}
are maximized by the same action $a \in (A_i)_s$. Here $\cC_{[i,n]}$ denotes the composite MDP of the truncated zig-zag diagram
\begin{align*}
        \xymatrix@R-=0.3cm@C-=0cm{
        & \nN_i \ar[dl] \ar[dr] & & \nN_{i+1} \ar[dl] \ar[dr] & & \ldots & & \nN_{n-1} \ar[dr] \ar[dl] \\
        \mM_i & & \mM_{i+1} & & \mM_{i+2} & \ldots & \mM_{n-1} & & \mM_n.
        }
\end{align*}
Monotonicity is a strong assumption that helps make a formal argument but can be relaxed. It is related to the notion that myopic solutions to the above maximization problems are globally optimal. The experiments considered in this section satisfy monotonicity. For a non-example, one can consider a moving agent on a grid having to come within a certain distance of two locations, which are an equal distance away from the agent's starting point.

In practice, a zig-zag diagram can always be made forward-moving by removing the actions of $\nN_i$ that can potentially move the agent off $\nN_i$ back into $\mM_i$. This can be formalized as puncturing $\mM_i$ along the complement of $\nN_i$ and intersecting the result with $\nN_i$. The details are of general interest but not immediately relevant to the present paper, so we skip a further discussion.

\begin{restatable}{theorem}{policy} \label{Theorem 3}
Suppose that the zig-zag diagram~\eqref{zig-zag} is forward-moving and the optimal value function of $\cC_n$ is monotonic. Then, following the individually calculated policies $\pi_i$ on each component $\mM_i$ gives an optimal policy on the composite MDP $\cC_n$.
\end{restatable}

The denotational theory of zig-zag diagrams in RL enables a structured, abstract, and rigorous understanding of complex sequential tasks via the following properties.
\begin{itemize}
\item \textbf{Semantics:} 
Denotational languages map syntactic constructs to mathematical objects, ensuring a precise understanding of what each part of a system means. Zig-zag diagrams represent the relationships between MDPs and subprocesses in a sequential task and encode the semantics of how these processes interact.
\item \textbf{Compositionality:} 
The meaning of the constituent parts determines the meaning of a complex expression, ensuring modularity. Similarly, the zig-zag pattern shows how complex processes comprise smaller subprocesses and individual MDPs.
\item \textbf{Abstraction:} 
Denotational languages abstract away many implementation details and focus on the meaning or behavior of constructs. Zig-zag diagrams abstract away the specific workings of each MDP and subprocess, focusing instead on their high-level relationships.
\item \textbf{Formal system:} The relationships in a zig-zag diagram are subject to specific mathematical conditions and definitions, providing a formal and rigorous understanding of the system.
\end{itemize}

The compositional theory based on categorical operators is not constrained to sequential scenarios. It is applicable to any decomposition into sub-tasks, including cyclical patterns (which we see as a feature). We chose to treat zig-zag diagrams because we can practically model realistic system scenarios and theoretically guarantee optimality for sequential task composition.

\begin{figure}[!t]
    \centering
    \includegraphics[clip, trim=9cm 5cm 9cm 2.5cm, width=0.48\linewidth]{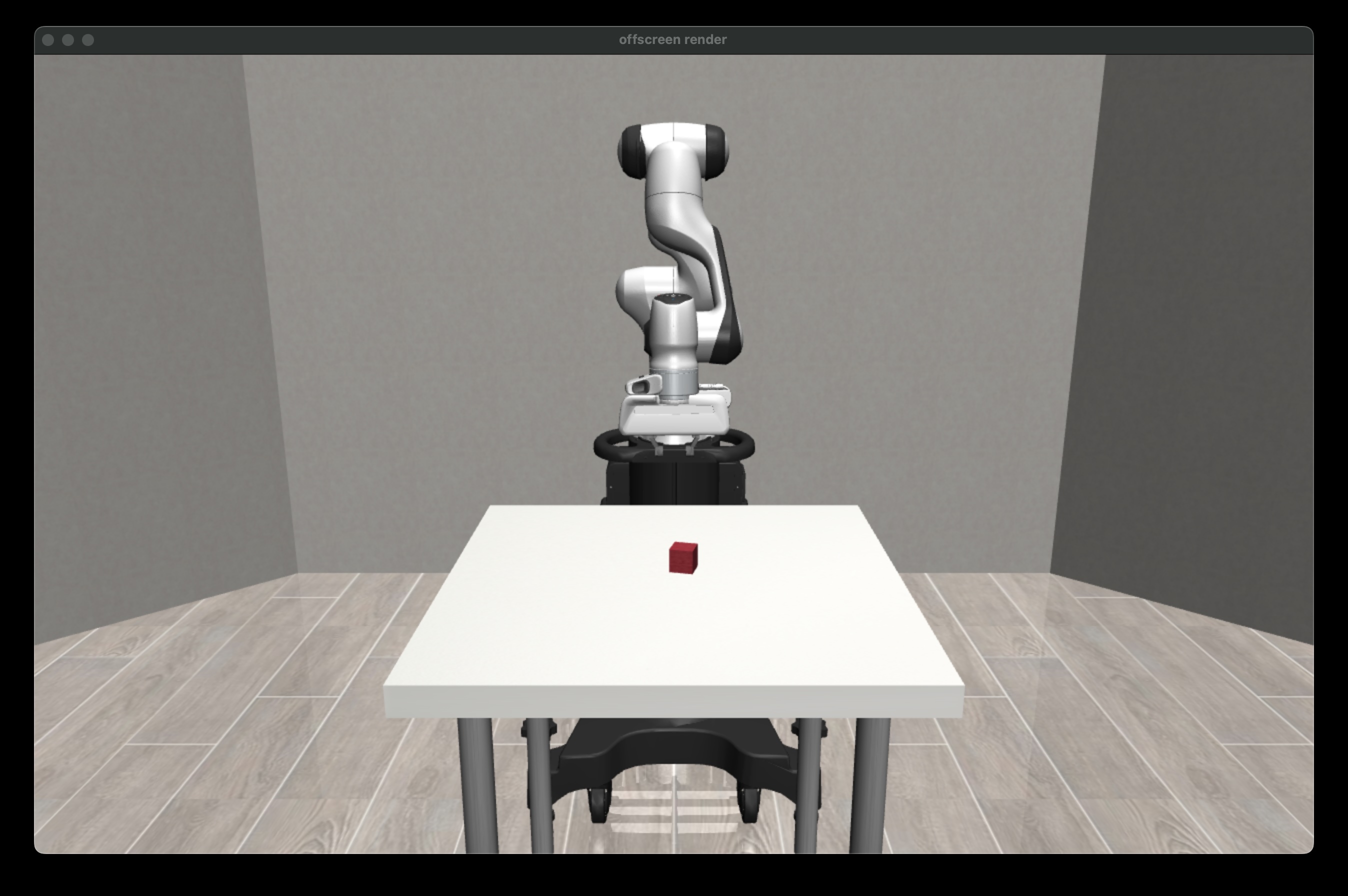}
    \includegraphics[clip, trim=9cm 5cm 9cm 2.5cm, width=0.48\linewidth]{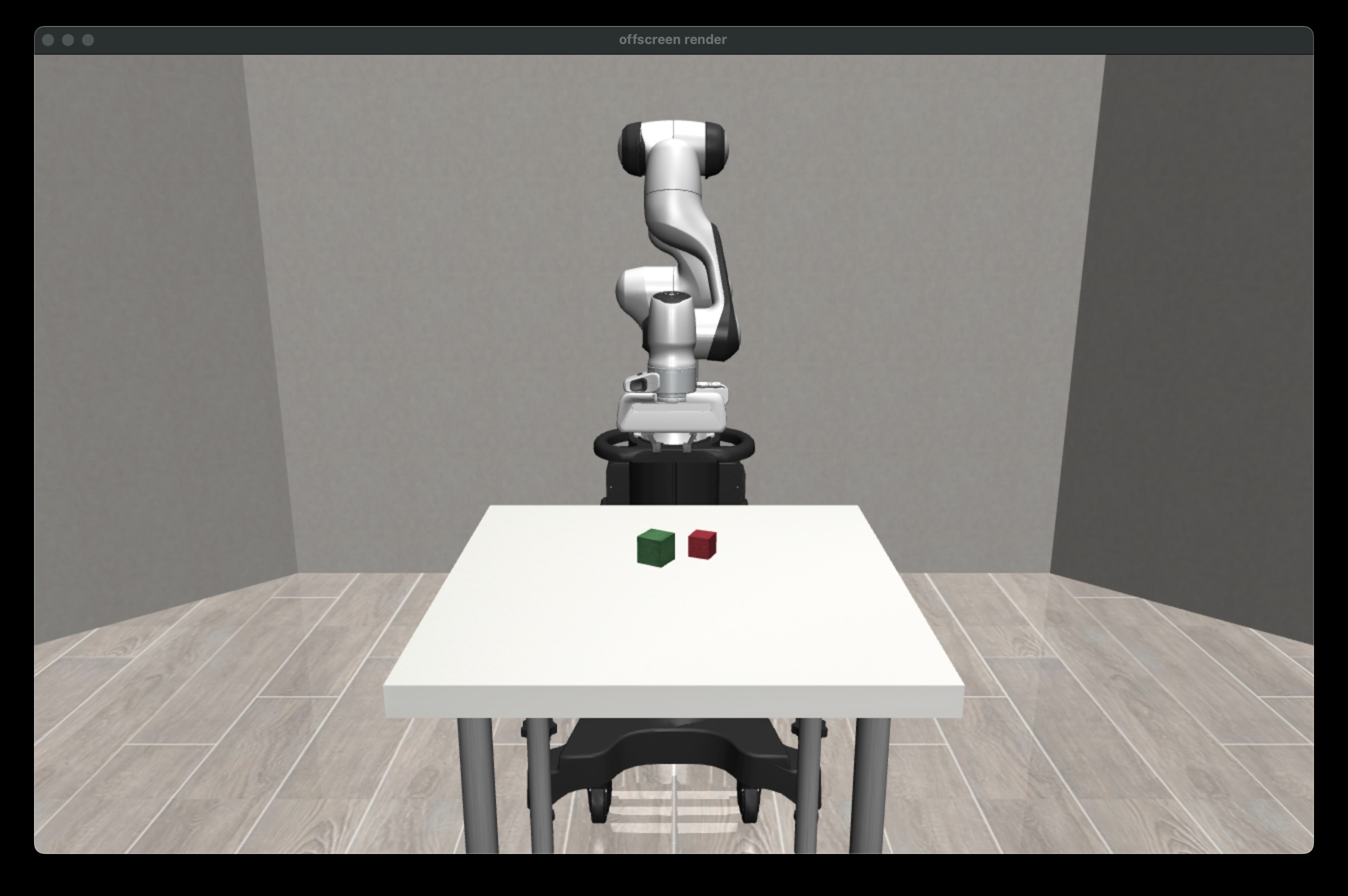}\\
    \includegraphics[clip, trim=9cm 5cm 9cm 2.5cm, width=0.48\linewidth]{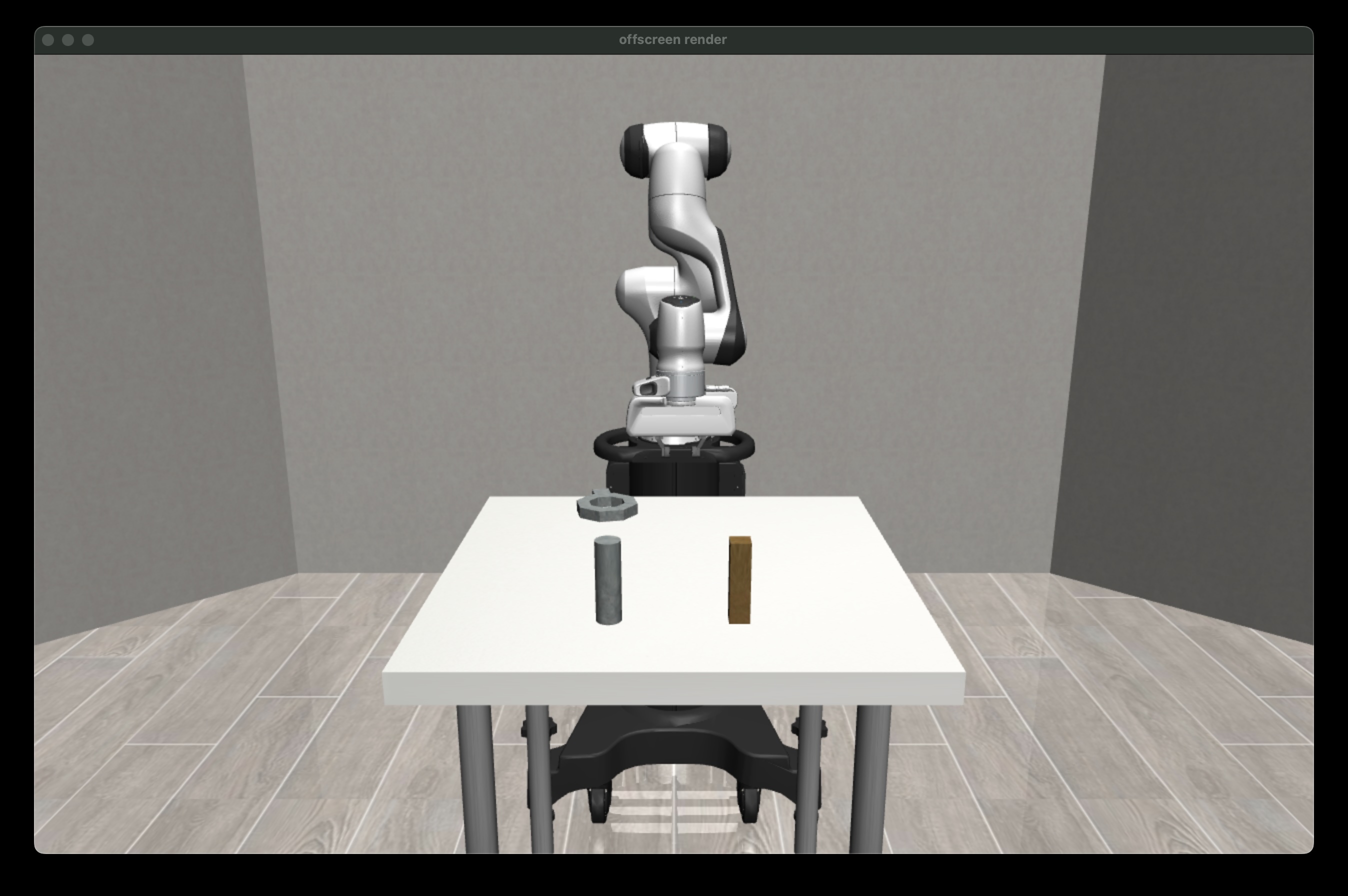}
    \includegraphics[clip, trim=9cm 5cm 9cm 2.5cm, width=0.48\linewidth]{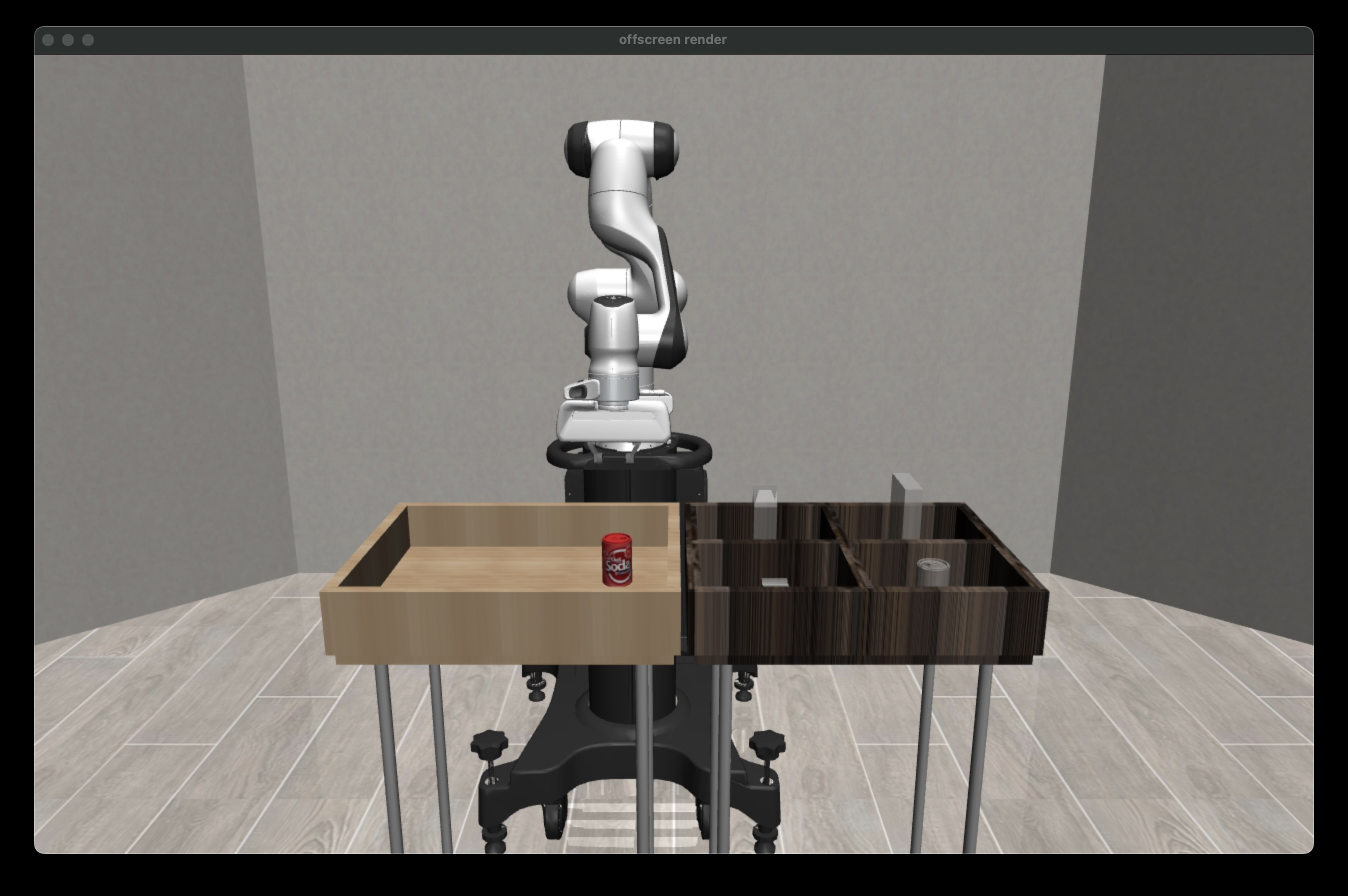}
    \caption{Category-theoretic compositional RL achieves efficient solutions for increasingly complex tasks.}
    \label{fig:results-sr1}
    \vspace{2em}
\end{figure}

\subsection{Experiments}

To support the categorical formalism through experiments, we tasked RL agents with four distinct manipulation challenges (figure~\ref{fig:results-sr1}) using robosuite~\citep{robosuite:2020, bakirtzis:2024}.
\begin{enumerate}[label=Task \arabic*,leftmargin=*]
\item \emph{Lift a block} \; The robot lifts a single block:
\begin{align*}
        \xymatrix@R-=0.3cm@C-=0cm{
        & \mathsf{reached} \ar[dl] \ar[dr] & & \\
        \mathsf{reach} & & \mathsf{lift}.
        }
\end{align*}
\item \emph{Stack blocks}  \; The robot stacks one block on top of another:
\begin{align*}
        \xymatrix@R-=0.3cm@C-=0cm{
        & \mathsf{reached} \ar[dl] \ar[dr] & & \mathsf{raised} \ar[dl] \ar[dr] & & \\
        \mathsf{reach} & & \mathsf{lift} & & \mathsf{place}.
        }
\end{align*}
\item \emph{Assemble a nut} \; The robot picks up a nut from the table top and fits it around a pole:
\begin{align*}
        \xymatrix@R-=0.3cm@C-=0cm{
        & \mathsf{reached} \ar[dl] \ar[dr] & & \mathsf{raised} \ar[dl] \ar[dr] & & \\
        \mathsf{reach} & & \mathsf{lift} & & \mathsf{place}.
        }
\end{align*}
\item \emph{Pick and place a can} \; The robot retrieves a soda can and places it in a designated bin:
\begin{align*}
        \xymatrix@R-=0.3cm@C-=0cm{
        & \mathsf{reached} \ar[dl] \ar[dr] & & \mathsf{raised} \ar[dl] \ar[dr] & & \mathsf{crossed} \ar[dl] \ar[dr] & & \\
        \mathsf{reach} & & \mathsf{lift} & & \mathsf{transport} & & \mathsf{place}.
        }
\end{align*}
\end{enumerate}
The experiments within the robosuite simulator demonstrate a compositional approach to RL by integrating category theory. This integration emphasizes the use of zig-zag diagrams as an efficient denotational tool. The resulting compositional RL technique provides a direct method for sequential decision-making and introduces modularity in robotic tasks, adding precision to learning.

The practical strengths of mapping computational tasks to mathematical objects manifest in the following properties.
\begin{itemize}
    \item \textbf{Compositionality:} When tasks break down into sub-tasks, the semantics guarantee that the entirety's meaning is an aggregation of its components, streamlining the synthesis of intricate tasks from foundational ones.
    \item \textbf{Scalability:} Scaling up and integrating new computational tasks becomes unambiguous.
    \item \textbf{Interoperability:} Employing a common mathematical framework ensures consistent and modular understanding across varied system compositions.
\end{itemize}

\subsubsection{Zig-zag Task Composition and Reward Structure}

In the context of the zig-zag diagrams and the categorical formalism, each manipulation task corresponds to an environment $\mM_i$, and its series of sub-tasks align with the subprocesses $\nN_i$. We have a defined dense reward signal $r_\textrm{dense}$ and a set success criterion for every such sub-task. Meeting this success criterion implies the agent has reached a state within the subprocess $\nN_i$ and receives a task reward $r_\textrm{task}$. However, if the agent meets the failure criterion, it signifies a deviation from the intended subprocess path, resetting the agent to the beginning of the task or environment $\mM_0$.

Each sub-task within an environment $\mM_i$ is associated with a subprocess $\nN_i$, and we dedicate an individual RL agent to each such subprocess. During training, the agent corresponding to the active subprocess or sub-task samples an action records the subsequent experience, and refines its policy. From the perspective of the categorical structure, this approach resembles traversing through the zig-zag diagrams sequentially. Initially, with all agents set to random policies, the training effectively starts with the environment $\mM_0$ and its associated subprocess $\nN_0$. Only after achieving success in this initial sub-task does the robot begin accumulating experiences in the subsequent subprocesses or sub-tasks, moving through the diagram.

Here are the settings for each sub-task MDP mapping to the zig-zag diagrams above.
\begin{itemize}
    \item $\mM_{\mathrm{r}}[r, o]$ models the reaching task, where the robot $r$ must reach its grippers around an object $o$ of interest. The 6-dimensional state space consists of the position of the robot hand and the object: $\mathbf{s} = [\mathbf{p}_r, \mathbf{p}_o]$. At each step, the robot controls the displacement of its hand, moving at most 10 cm in each direction: $\mathbf{a} = \Delta \mathbf{p}_r$. When the MDP transitions from state $\mathbf{s}$ to $\mathbf{s'}$, the robot receives a shaped reward $r = ||\mathbf{p}_r - \mathbf{p}_o||_2 - ||\mathbf{p'}_r - \mathbf{p'}_o||_2$.
    \item $\mM_{\mathrm{l}}[r, o]$ models the lifting task, assuming the robot hand is aligned with the object. The robot is only allowed up-and-down movements plus gripper controls. The state space contains the robot hand height, object height, and gripper width: $\mathbf{s} = [z_r, z_o, w]$, and $\mathbf{a} = [\Delta z_r, \Delta w]$. The robot is rewarded for staying close to the object and lifting it off the table: $r = (|z_r - z_o| - |z'_r - z'_o|) + (z'_r - z_r)$.
    \item $\mM_{\mathrm{t}}[r, o]$ is the transporting task. The robot must transport a held object to the other side of the table. The robot controls the hand movement but cannot open the gripper. Because the robot is holding the object, they share the same position. So $\mathbf{s} = \mathbf{p}_r$, and $\mathbf{a} = \Delta \mathbf{p}_r$. The robot is rewarded for moving in the $+Y$ direction: $r = y'_r - y_r$.
    \item $\mM_{\mathrm{p}}[r, o, g]$ is the placing task. Starting with the robot holding the object, it must place the object at the desired goal location $g$. The state space contains the robot position, object position, goal position, and gripper width: $\mathbf{s} = [\mathbf{p}_r, \mathbf{p}_o, \mathbf{p}_g, w]$. The robot controls the hand movement and gripper opening: $\mathbf{a} = [\Delta \mathbf{p}_r, \Delta w]$.
\end{itemize}

\begin{figure}[!t]
    \centering
    \includegraphics[clip, trim=0.5cm 0.5cm 0.5cm 0.5cm, width=0.49\linewidth]{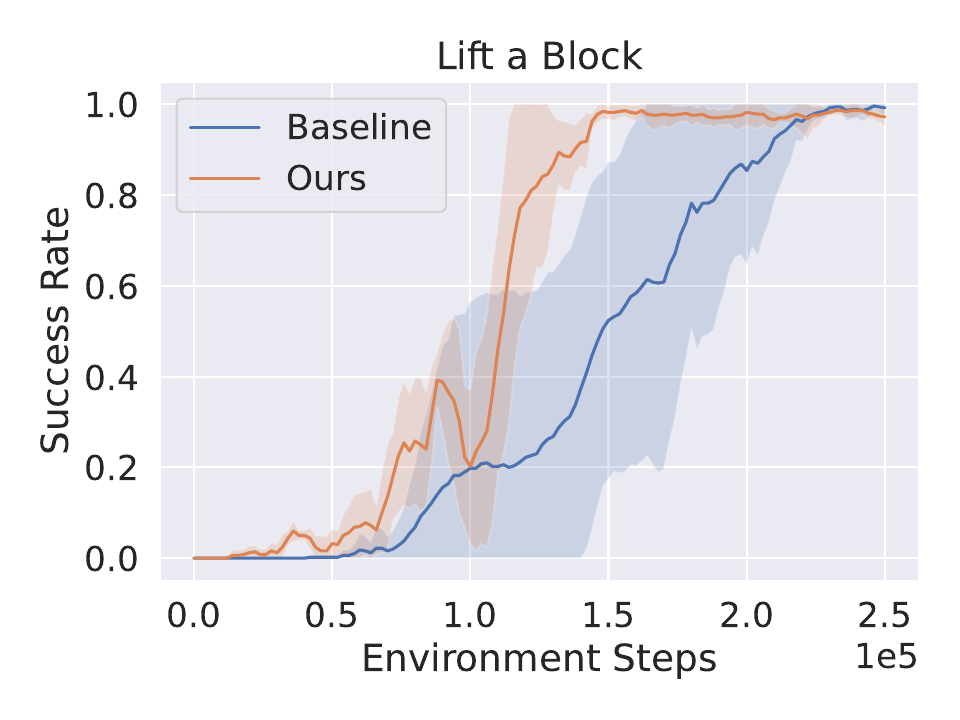}
    \includegraphics[clip, trim=0.5cm 0.5cm 0.5cm 0.5cm, width=0.49\linewidth]{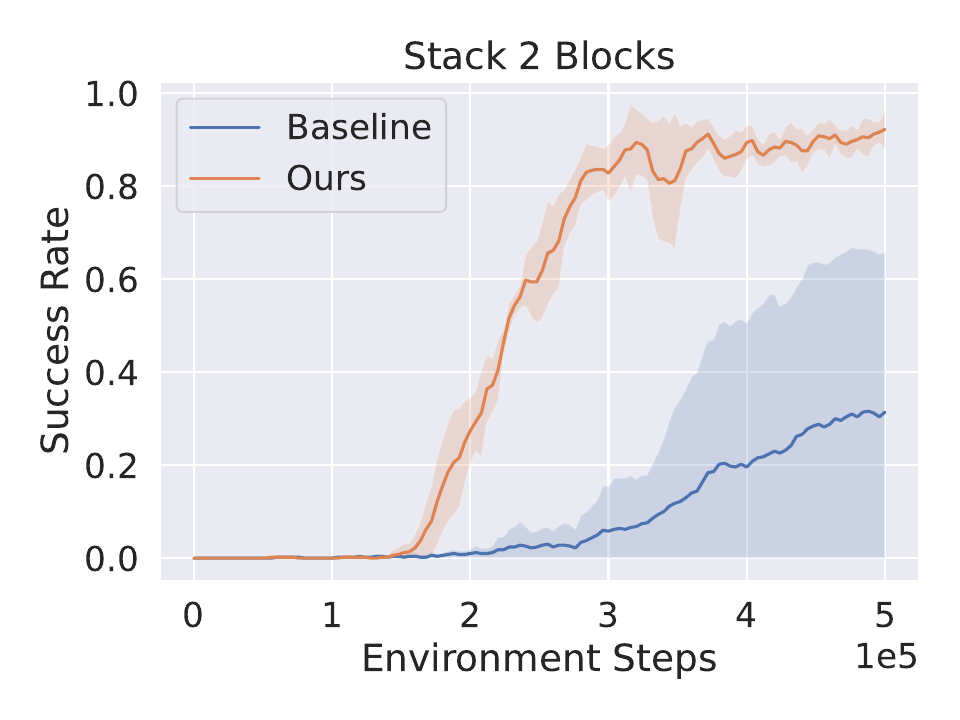}\\
    \includegraphics[clip, trim=0.5cm 0.5cm 0.5cm 0.5cm, width=0.49\linewidth]{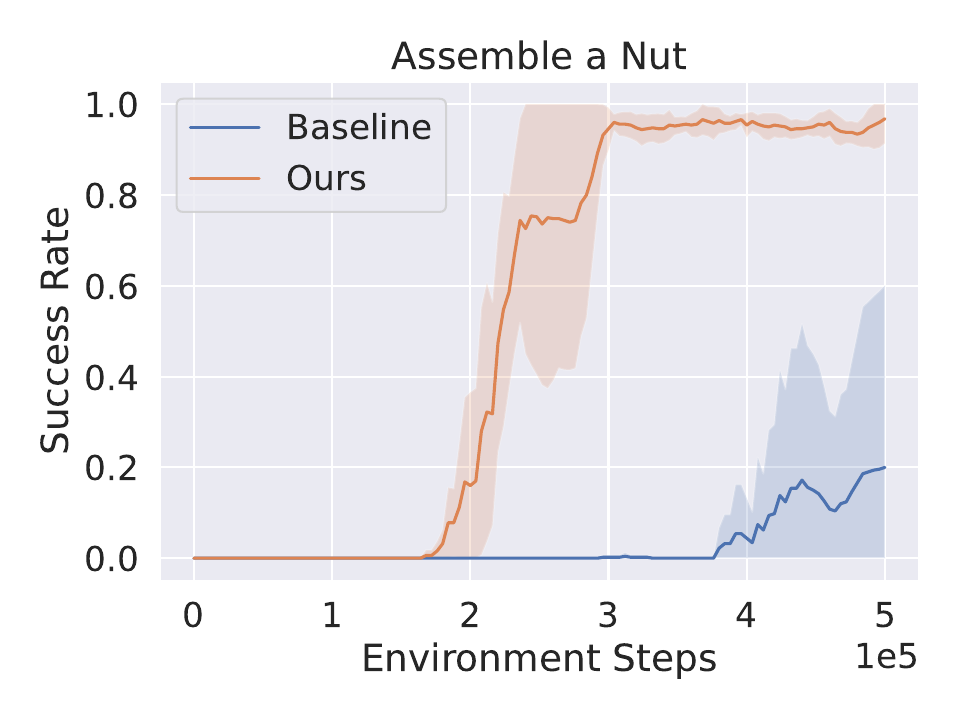}
    \includegraphics[clip, trim=0.5cm 0.5cm 0.5cm 0.5cm, width=0.49\linewidth]{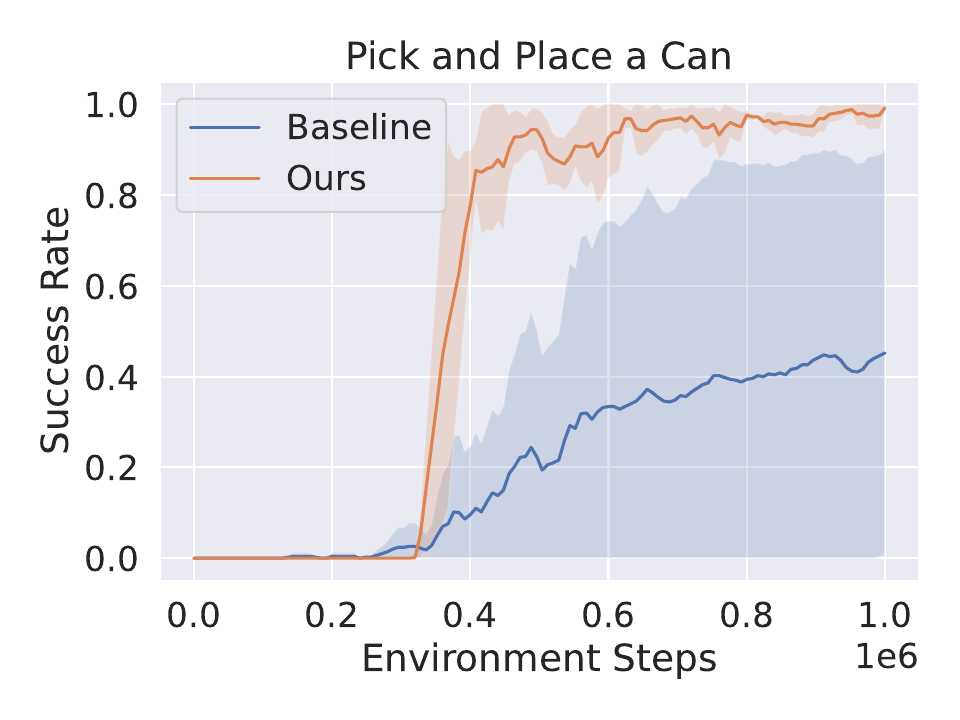}
    \caption{Category-theoretic compositional RL vs. baseline MDP: A 50\% gain in sample efficiency for block-lifting, with demonstrated capability in complex tasks.}
    \label{fig:results-sr}
    \vspace{2em}
\end{figure}

Next, we define the subprocesses corresponding to the sub-task MDPs.
\begin{itemize}
    \item $\nN_{\mathrm{r}}[r, o]$ contains the states where the robot position is at most 1 cm away from the object: $||\mathbf{p}_r - \mathbf{p}_o||_2 < 0.01$.
    \item $\nN_{\mathrm{l}}[r, o]$ is entered when the object's height is above some threshold $h$: $y_o > h$.
    \item $\nN_{\mathrm{t}}[r, o]$ contains the states where the robot has crossed the center line of the table: $y_r > 0.03$.
\end{itemize}

We use the same reward structure for the baseline. In particular, the baseline to compare against trains a single RL agent to complete the total task, and the reward given to the agent is the summation of the rewards given to the zig-zag diagram. Additionally, robosuite incorporates built-in stochasticity regarding objects' initial locations for all experiments. The success rate is computed from $20$ evaluation episodes with stochastic object placements. We perform evaluations every $2000$ training steps.

\subsubsection{Performance Evaluation of Subprocess Composition}

Decomposing complex, long-horizon manipulation tasks into smaller subprocesses enhances learning performance, as evidenced by success rate and convergence speed improvements. This improvement is captured through the formalism of zig-zag diagrams, which provide a structured approach to understanding the relationship between sub-tasks within the larger task. In this framework, the zig-zag diagrams represent the sub-tasks composition, connecting states and transitions within the MDP. By systematically breaking down complex tasks, we enable more efficient learning and synthesis of solutions, leading to observed performance improvements. 

\begin{figure}[!t]
    \centering
    \includegraphics[clip, trim=0.5cm 0.5cm 0.5cm 0.5cm, width=0.49\linewidth]{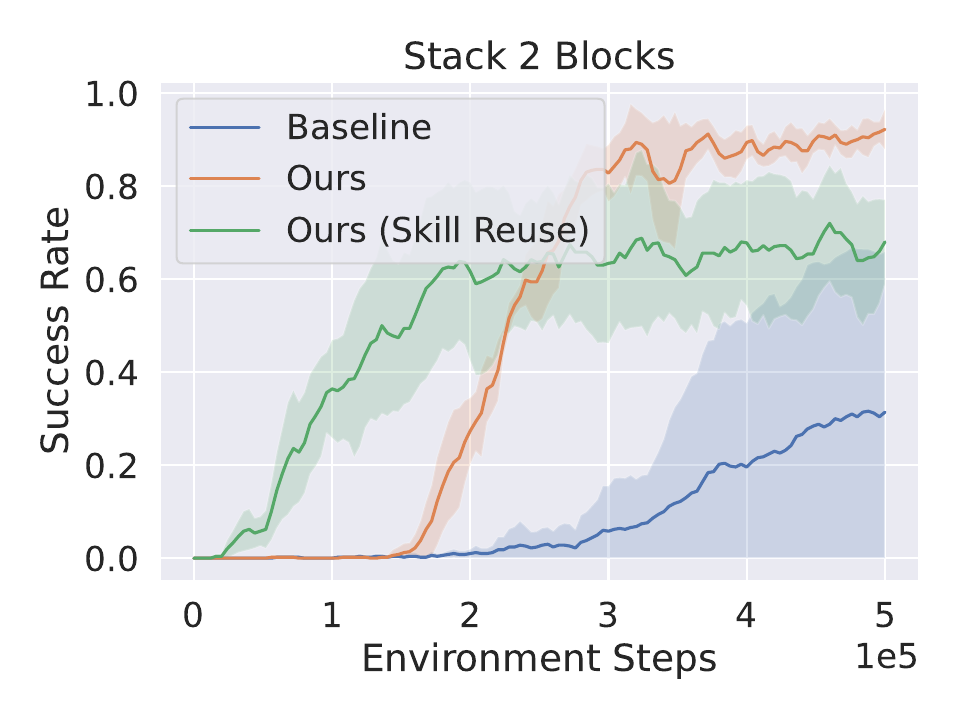}
    \includegraphics[clip, trim=0.5cm 0.5cm 0.5cm 0.5cm, width=0.49\linewidth]{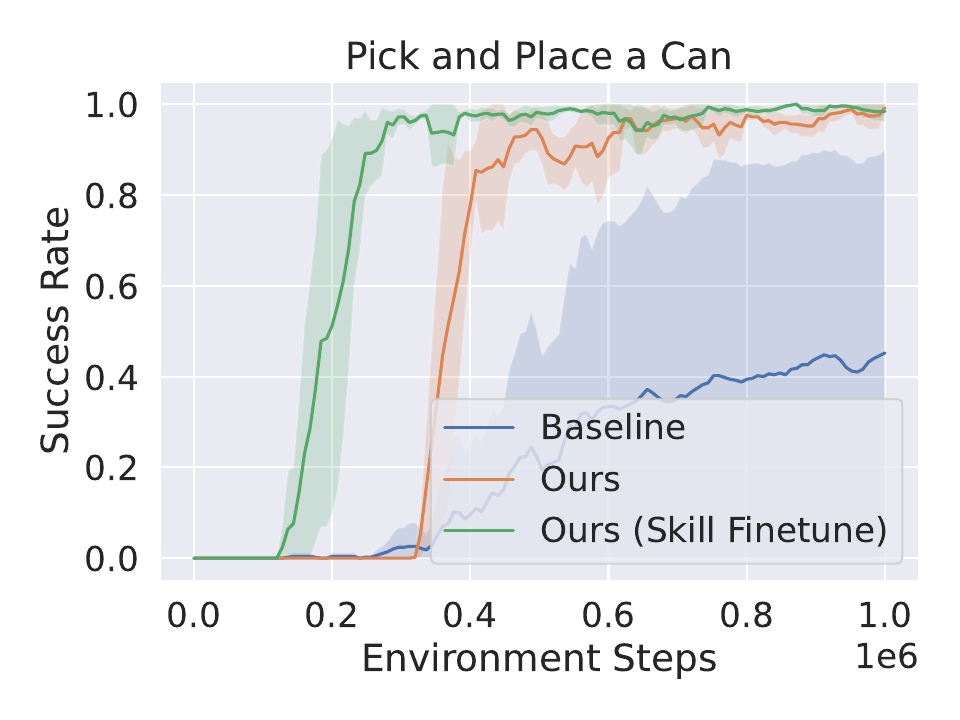}
    \caption{Compositional RL enables reusing and recycling sub-task policies from previously learned tasks, improving sample efficiency.}
    \label{fig:results-reuse-recycle}
    \vspace{2em}
\end{figure}

We designed a robosuite environment wrapper using categorical constructs to oversee and transition between sub-tasks. This wrapper filters state vectors at each step, modifies actions per the defined sub-task MDPs, and allocates dense rewards $r_\textrm{dense}$. Completing a sub-task triggers a $r_\textrm{task} = 10$ reward and transitions to the subsequent sub-task. Upon task completion or environment termination, it reverts to the initial sub-task. This approach of breaking a long-horizon control task into shorter segments facilitates training individual RL agents for each segment. For comparison, we train using a direct robosuite environment with a similarly constructed dense reward. Both methods use soft actor-critic \citep{haarnoja:2018}, an advanced model-free RL algorithm.

We conduct experiments using $5$ random seeds for each setting. The category-theoretic compositional RL performs well in training sample efficiency and final model performance (figure~\ref{fig:results-sr}). In the block-lifting task, our method converges to a $100$\% success rate after $150$k training steps, whereas the baseline method converges at around $225$k steps. In a more challenging task like block-stacking, our method converges to over $90$\% success rate while the baseline method struggles to reach even a $50$\% success rate. The trend continues to the nut-assembly and can-moving tasks where our method consistently learns a better policy with fewer training steps.

Task composition enables the reuse or recycling of existing trained sub-task policies. Because all four robosuite tasks involve the reach and lift sub-tasks, repetitive training can be avoided after training the block-lifting task. Aside from direct reuse, where the trained policies are directly used in the sub-task of a different environment, recycling could prove beneficial: the interactions and reward signals from the new sub-task are used to fine-tune the policy parameters. Reusing the reach and lift skills allows our method to start training directly from the place sub-task, significantly improving sample efficiency (figure~\ref{fig:results-reuse-recycle}). 
However, the final performance is lower than when compositional RL is trained from scratch. When a second block is present, the policies trained to lift a single block occasionally fail due to the robot hand getting stuck on the other block. When further fine-tuning is performed, the lifting policies initially trained on the block quickly adapt to the soda can with a different size.

We achieve higher success rates and faster convergence by breaking down complex, long-horizon tasks into systematic sub-tasks, as represented through zig-zag diagrams. Moreover, the reuse and recycling of trained sub-task policies highlight the adaptability of zig-zag diagrams, avoiding redundancy and further boosting efficiency. The experimental results reinforce that categorical formalism is a robust foundation for composition and structure in RL.

\subsubsection{Limitations} 

A comprehensive compositional generalization benchmark is still lacking~\cite{mendez:2022a,gur:2021}. This absence hinders the ability to systematically evaluate and compare the performance of different compositional RL algorithms across a standardized set of tasks. If such benchmarks existed, it would be straightforward to determine which algorithms are more effective at generalizing from their training environments to unseen scenarios. Furthermore, the need for standardized evaluation metrics for compositional generalization in RL adds another layer of complexity. Metrics that can accurately reflect the ability of an algorithm to leverage compositional structures in learning and decision-making processes are essential for advancing the field. This methodological gap means that current algorithm comparisons rely on inconsistent criteria or non-comparable tasks. We attempted to provide a common baseline that does not compare sparse reward structures with unfair dense reward structures. Future directions will attempt to derive a fair benchmark for compositional RL algorithms and compare compositional RL algorithms with our proposal based on categorical structures.

\section{Related Work}

Our work builds upon and extends a rich tradition in RL that explores the structure and abstraction in decision processes~\cite{lavaei:2023,ravindran:2003,ravindran:2004} and recent developments in applied category theory~\cite{aguinaldo:2022,bakirtzis:2021c,bakirtzis:2021a, bakirtzis:2021b,cruttwell:2022,hanks:2024,hedges:2024,zardini:2021a,zardini:2021,zardini:2023}. Seminal contributions on minimalization have laid foundational concepts for understanding state and action abstractions in MDPs \citep{givan:2003}. Similarly, the work on factored and propositional representations has been pivotal in advancing structured solution techniques and symbolic dynamic programming \citep{boutilier:1995,boutilier:2000}. Additionally, work on the theoretical underpinnings of structured solution techniques in RL provides a way to synthesize behaviors \citep{otterlo:2009}.

In hierarchical reinforcement learning (HRL), significant formal models have been developed that go beyond the heuristic layering of policies. These models provide structured approaches to defining subtasks and subgoals, facilitating multi-task learning and systematic problem decomposition \citep{parr:1997,dietterich:2000,ravindran:2013,nachum:2018}. Our categorical approach aims to integrate these hierarchical structures within a unified mathematical framework, offering a complementary perspective on task composition and policy integration.

Building on foundational principles, compositionality in RL has traditionally focused on temporal and state abstractions to execute complex behaviors and enhance learning efficiency through mechanisms such as skill chaining \citep{tasse:2020,niekerk:2019,jothimurugan:2021,ivanov:2021,mendez:2022a}. In contrast, our categorical formalism introduces a shift towards functional composition in RL. This approach leverages the denotational nature of category theory to decompose tasks into distinct behavioral functions, providing a more structured and mathematically rigorous framework for task decomposition than previously available.

This functional approach offers granularity and aligns with contemporary efforts in robotics and policy modularization. For instance, it complements methods used in robotics tasks \citep{devin:2017}, where decomposing complex behaviors into simpler, manageable units is crucial. Similarly, it supports the development of modular neural architectures for policy learning \citep{mendez:2022}, where each module can be understood and optimized independently. Our framework enriches these efforts by providing a formal language of zig-zag diagrams, which aids in the systematic composition and decomposition of tasks and policies. This diagrammatic language not only enhances the interpretability of complex decision-making structures but also ensures that these structures can be rigorously analyzed and validated within a coherent theoretical framework.

Our categorical formalism addresses the need for a unified and rigorous mathematical framework that can encapsulate and generalize the concepts of MDP homomorphisms and task composition. Unlike previous works, our approach leverages the  semantics of category theory to provide a systematic structure for decomposing and recomposing tasks and policies. This is achieved through the introduction of categorical operations such as pushouts, which we prove exist for MDPs and serve as a novel method for task integration.

We extend the work on probabilistic representations in category theory, moving beyond the stochastic process descriptions~\citep{giry:1981,watanabe:2023} and recent categorical treatments of MDPs~\citep{baez:2016,fritz:2023}. Our framework not only models the dynamics of decision processes but also provides a compositional toolset for functional decomposition in RL, which has been less explored in the existing literature.

In this paper, we examine how the categorical approach can enhance RL systems' modularity, scalability, and interoperability. By providing a rigorous mathematical structure for task and policy composition, our approach offers potential improvements in learning efficiency and adaptability in complex environments. However, we also acknowledge the challenges and limitations of applying abstract mathematical frameworks in practical RL scenarios.

\section{Conclusion}

In this work, we introduce an abstraction theory for compositional RL, underpinned by the formal construct of categorical pushouts in the context of MDPs. Our adoption of zig-zag diagrams is a structured way to visualize and analyze the complex interdependencies between tasks and subprocesses within RL systems. Zig-zag diagrams are but one way to describe and synthesize sequential decision-making problems.

The categorical formalism we propose offers a denotational language that aids in precisely modeling decision-making scenarios in general, robotic tasks being only one of the application domains. The empirical results from our experiments indicate that this approach can lead to improvements in learning precision and sample efficiency. However, it is essential to mention that these findings are contextual and derived from specific task settings in controlled environments.

As we look to the future, the application of category theory in RL presents a promising avenue for research with the potential to address several open problems in compositional RL: task synthesis, generalization, and interpretability, to name a few. However, its broader impact and utility remain to be fully explored. We anticipate that further studies will be necessary to validate the scalability of this approach across diverse and more complex scenarios. Our ongoing research will refine this framework's theoretical tools and constructs to enhance their robustness and applicability.

\section{Acknowledgments}

G.B. and U.T. would like to acknowledge their support from the AFOSR FA9550-19-1-0005 grant.

\bibliography{manuscript}

\end{document}